\documentclass[twoside,11pt]{article}

\usepackage{jmlr2e}

\usepackage{booktabs}       % professional-quality tables
\usepackage{amsfonts}       % blackboard math symbols
\usepackage{nicefrac}       % compact symbols for 1/2, etc.
\usepackage{tabularx}
\usepackage{multicol}
\usepackage{rotating}
\usepackage{multirow}
\usepackage{amsmath}

\usepackage{xcolor}
\ShortHeadings{ADAPT : Awesome Domain Adaptation Python Toolbox}{de Mathelin et al.}
\firstpageno{1}

\begin{document}

\title{ADAPT : Awesome Domain Adaptation Python Toolbox}

\author{\name Antoine de Mathelin$^{1, 2}$ \email antoine.de-mathelin-de-papigny@michelin.com \\
       \name Mounir Atiq$^{2}$ \email mounir.atiq@ens-paris-saclay.fr \\
       \name Guillaume Richard$^{2}$ \email guillaume.richard@ens-paris-saclay.fr \\
       \name Alejandro de la Concha$^{2}$ \email alejandro.de\_la\_concha\_duarte@ens-paris-saclay.fr \\
       \name Mouad Yachouti$^{2}$ \email mouad.yachouti@ens-paris-saclay.fr \\
       \name François Deheeger$^{1}$ \email francois.deheeger@michelin.com \\
       \name Mathilde Mougeot$^{2}$ \email mathilde.mougeot@ens-paris-saclay.fr \\
       \name Nicolas Vayatis$^{2}$ \email nicolas.vayatis@ens-paris-saclay.fr \\
       \addr $^{1}$Manufacture Française des pneumatiques Michelin,
       Clermont-Ferrand, 63000, France \\
       \addr $^{2}$Centre Borelli, Université Paris-Saclay, CNRS, ENS Paris-Saclay, Gif-sur-Yvette, 91190, France
       }

% \editor{Kevin Murphy and Bernhard Sch{\"o}lkopf}

\maketitle

\thispagestyle{plain}

\begin{abstract}

In this paper, we introduce the ADAPT library, an open source Python API providing the implementation of the main transfer learning and domain adaptation methods. The library is designed with a user friendly approach to facilitate the access to domain adaptation for a wide public. ADAPT is compatible with scikit-learn and TensorFlow and a full documentation is proposed online \url{https://adapt-python.github.io/adapt/} with a substantial gallery of examples.

% ADAPT is an open-source python library providing the implementation of the main transfer learning and domain adaptation methods. The purpose of this. A full documentation is proposed online \url{https://adapt-python.github.io/adapt/} with a substantial gallery of examples.

% The library offers three modules corresponding to the three principal strategies of domain adaptation: (i) \textit{feature-based} containing methods performing feature transformation; (ii) \textit{instance-based} with the implementation of reweighting techniques and (iii) \textit{parameter-based} proposing methods to adapt pre-trained models to novel observations. A full documentation is proposed online \url{https://adapt-python.github.io/adapt/} with gallery of examples. Besides, the library presents an high test coverage.
\end{abstract}


\section{Introduction and Motivation}

% Over the past few years, many domain adaptation techniques have been proposed to improve machine leanring model performances.

% The need of domain adaptation raise from the 
% In recent years, many domain adaptation methods have been developed in order to improve machine learning model performances.

% Transfer learning and domain adaptation refer to the field of machine learning where the learner 

% In traditional machine learning scenarios, the learner assumes that the training data are drawn according to a source distribution identical to the target distribution on which the learned model is applied. However, in most practical scenarios, a "domain shift" exists between the source and the target distribution. To correct this shift, a large variety of domain adaptation techniques have been developed recently with various approaches as kernel-based optimization \citep{Huang2007KMM, Sugiyama2007KLIEP}, decision trees and ensemble models \citep{Segev2017SERSTRUCT, Pardoe2010boost}, deep learning \citep{Ganin2016DANN, Zhang2019MDD} etc...

% Domain adaptation methods are useful for application where the target distribution differs from the training distribution.

Transfer learning and domain adaptation (DA) aim to correct the shifts that exist between the training distribution of a machine learning model (referred as source) and the target distribution on which the model is deployed. This research field has known an important development over the past decades. The presence of "domain shifts" between source and target distributions is the typical framework encountered in real-life applications which makes domain adaptation algorithms particularly usefull for numerous use-cases. Domain adaptation techniques are often needed in scenarios where labels are easily available on a source domain but are expensive on the target domain. For example, it may be used to leverage information from a synthetic dataset to build a classifier for real data \citep{Ganin2016DANN} such as the adaptation of GTA images for autonomous car segmentation \citep{Saito2018MCD}. Furthermore, domain adaptation is of great interest to correct bias from the training samples such as sample bias \citep{Huang2007KMM} when one population in the training set is over-represented with respect to the overall population. DA methods are also used to correct shifts in the input features, caused by sensor or technological drifts \citep{Courty2016OTDA}. Finally domain adaptation is useful for specifying a pre-trained model on a sub-task with few labels, as segmentation of medical images \citep{Ravishankar2016TLMedicalImages}.

% the machine learning community is requiring an easy and fast access to several DA techniques implemented on the same basis. Motivated by real world applications,

Nowadays, domain adaptation has become an essential tool to handle domain shifts in real applications. Many DA methods have been developed in recent years and a large number of DA implementations are spread on the web. Paradoxically, the large number of DA variants makes their accessibility to users more difficult. Indeed, finding which methods will best fit a given domain adaptation problem is a difficult task. In practice, the user would like to try different methods and select the most promising one. However, most DA algorithms available on the web are not implemented under the same basis, some of them rely on PyTorch \citep{Paszke2019Pytorch}, others use scikit-learn \citep{Pedregosa2011scikit-learn} or TensorFlow \citep{Abadi2015Tensorflow}. Moreover, most open source implementations have for primary objective to reproduce the experimental results of a particular publication and an extra effort is then required to apply them on other data. Facing these difficulties, we propose ADAPT\footnote{\url{https://github.com/adapt-python/adapt}}, a new open-source Python library compatible with scikit-learn and TensorFlow. This library aims at facilitating the access to transfer learning and domain adaptation methods to a wide public including industrial practitioners. Inspired by the scikit-learn library, which manages to make machine learning accessible to everyone, ADAPT offers DA methods that can be easily used in the same format as each object implements the \textbf{\textit{fit}}, \textbf{\textit{predict}} and \textbf{\textit{score}} functions as any scikit-learn estimator. In a model deployment perspective, the ADAPT objects are compatible with the convenient features of scikit-learn as cloning and gridsearch. DA specific metrics are provided to allow an unsupervised selection of hyper-parameters. Finally, as we consider that the accessibility of a DA method essentially relies on the user understanding, a very detailed documentation is available online with many real and synthetic  examples. A user guide is provided to allow newcomers to find the right DA method for their problem based on practical considerations (see the ADAPT flowchart\footnote{\url{https://adapt-python.github.io/adapt/map.html}}). ADAPT offers the possibility of developing its own transfer method easily by subclassing existing classes. The library works on Linux, Mac and Windows for the four latest Python versions 3.6 to 3.9. ADAPT is nowadays appreciated by various users both from the academic and industrial world, with more than 1k downloads by month.

\section{Existing transfer libraries}

The emergence of domain adaptation algorithms first started around 2006. Progressively, many algorithms have been developed and open source implementations have been released. At some point, it became necessary to group several algorithms together to allow their comparison by the machine learning community. Some compilations of algorithms have then been developed as \textit{libTLDA} \citep{libTLDA} or \textit{DA-Toolbox} \citep{DAtoolbox}. These libraries implement "classic" DA methods, i.e which do not use deep learning, as KMM \citep{Huang2007KMM} or TCA \citep{Pan2010TCA}. These first attempts to group DA methods offered the opportunity to quickly test several methods on a same basis. However the proposed libraries were lacking of documentation and modularity. After that, many deep DA methods have been developed and some libraries have then proposed to group several variants under the same repository such as \textit{salad} \citep{Schneider2018salad}. Since then, many repositories for deep learning have been released, the most notable one being \textit{TLlib} \citep{Jiang2020dalib} which makes the great work of regrouping more than $40$ deep DA algorithms. It should be noted that all these repositories propose PyTorch implementations and are designed in a benchmark purpose, i.e. they mainly focus on comparing the results of each variant on well known datasets. Using these repositories on the user dataset often requires an extra effort.

% compared to calling a "\textit{fit}" method.

In 2021, ADAPT has been released, with the purpose to open the DA algorithms to newcomers and, in particular, industrial players. For this purpose, we group the main "classic" and "deep" DA methods into a same library and provide a very detailed documentation and robust code, tested through unit tests. The library is available on PyPI, offering more than 30 algorithms implemented in a user friendly scikit-learn style (cf Table \ref{comparison}).

\defcitealias{DomainBed}{Gulrajani et al. (2020)}
\defcitealias{pytorchada}{Tousch et al. (2020)}

\begin{table*}
	\label{table-all}
	\scriptsize
	\centering
	\begin{tabular}{c|c|c|c|c|c|c|c|c}
		\toprule
		Library & Reference & Eco  & PyPI & DDA & CDA & Doc & Test & Sk-Style \\
		\midrule
		Adapt & \cite{Demathelin2021ADAPT} & S+T & \checkmark & 12 & 22 & \checkmark & \checkmark & \checkmark  \\
		TL-Toolkit & \cite{Zhuang2019TransferToolkit} & S+T &  & 5 & 14 & & & \checkmark \\
		libTLDA & \cite{libTLDA} & S+M & \checkmark & 0 & 13 & \checkmark & \checkmark & \checkmark  \\
		TransferTools & \cite{transfertools} & S & \checkmark & 0 & 5 &  &  & \checkmark  \\
		DDAN & \cite{DDAN} & T & & 4 & 0 &  &  & \checkmark   \\
        TLlib & \cite{Jiang2020dalib} & P & \checkmark  & 40 & 0 & \checkmark &  &   \\
		DomainBed & \citetalias{DomainBed} & P & & 26 & 0 &  & \checkmark &   \\
		PyTorch-Adapt & \cite{PytorchAdapt} & P & \checkmark & 23 & 0 & \checkmark & \checkmark & \\
		Dassl & \cite{Dassl} & P & & 17 & 0 & & & \\
		salad & \cite{Schneider2018salad} & P & \checkmark & 11 & 0 & \checkmark & \checkmark &   \\
		Deep-TL & \cite{deepTL} & P & & 9 & 0 & & &  \\
		Pytorch-Ada & \citetalias{pytorchada} & P & \checkmark & 5 & 0 & \checkmark & \checkmark &   \\
		DA-Toolbox & \cite{DAtoolbox} & M &  & 0 & 10 &  &  &   \\
% 		Code ecosystem & scikit-learn, TensorFlow & PyTorch & PyTorch
		\bottomrule
	\end{tabular}
	\caption{DA repositories comparison. "Eco" refers to the code ecosystem, i.e. scikit-learn (S), TensorFlow (T), Pytorch (P) and Matlab (M). "DDA" and "CDA" give respectively the number of deep learning and classic DA algorithms in the library. Library implemented under the "Sklearn Style" (Sk-Style) are characterized by the presence of objects which implement "fit" and "predict" methods.}
	\label{comparison}
\end{table*}

\section{Organization}

The ADAPT library is divided into three modules : \textit{feature-based}, \textit{instance-based} and \textit{parameter-based} corresponding to the three main DA strategies. A list of all implemented methods is presented in Table \ref{table-all} (Appendix A). As mentioned in the DA survey \citep{Weiss2016survey}, some DA methods are based on the use of a small number of labeled target data and are referred as supervised domain adaptation methods (SDA), others use only unlabeled target data along with the sources (UDA). Some methods perform the adaptation and the learning of the task in one stage, others in two.

\subsection{Feature-Based Methods}

\begin{figure}[ht]
	\centering
	\includegraphics[width=\textwidth]{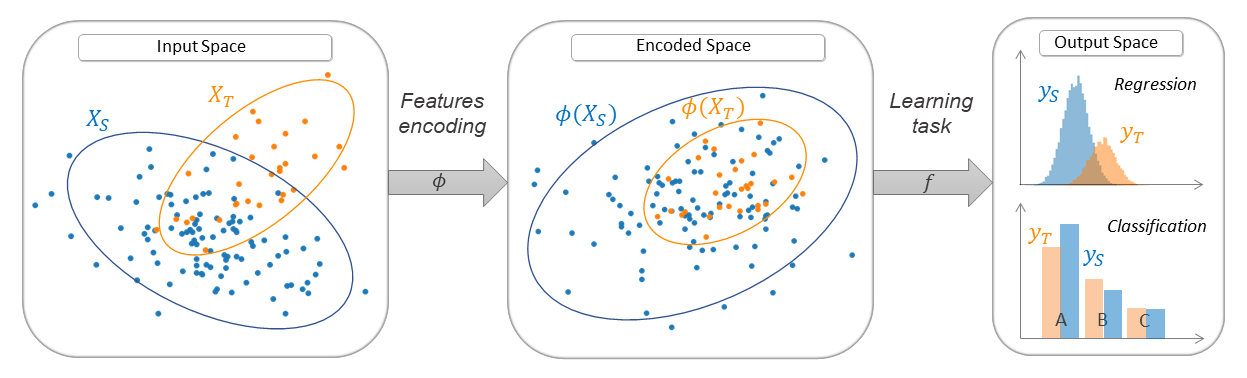}
	\caption{Feature-based strategy illustration.}
	\label{fig-feat}
\end{figure}

The purpose of feature-based DA methods (Figure \ref{fig-feat}) is to learn a new representation of the input features in which both source and target distribution match. This DA strategy is mostly used for unsupervised DA. Feature-based methods generally consider the assumption that the domain shift is due to an unknown transformation of the input space caused for instance by sensor drifts or any changes in the acquisition conditions \citep{Courty2016OTDA, Ganin2016DANN}.

\subsection{Instance-Based}

\begin{figure}[ht]
	\centering
	\includegraphics[width=\textwidth]{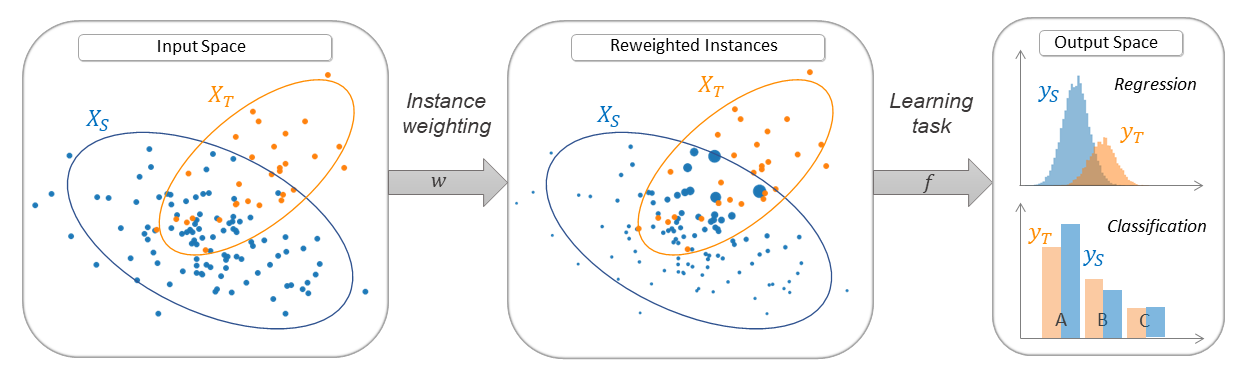}
	\caption{Instance-based strategy illustration}
	\label{fig-inst}
\end{figure}

The goal of instance-based methods (Figure \ref{fig-inst}) is to perform a reweighting of source instances in order to correct the difference between source and target distributions. This kind of methods are mostly used in \textit{sample bias} scenario and assume that source and target distribution share the same support in the input space \citep{Huang2007KMM, Sugiyama2007KLIEP}.

\subsection{Parameter-Based}

\begin{figure}[ht]
	\centering
	\includegraphics[width=\textwidth]{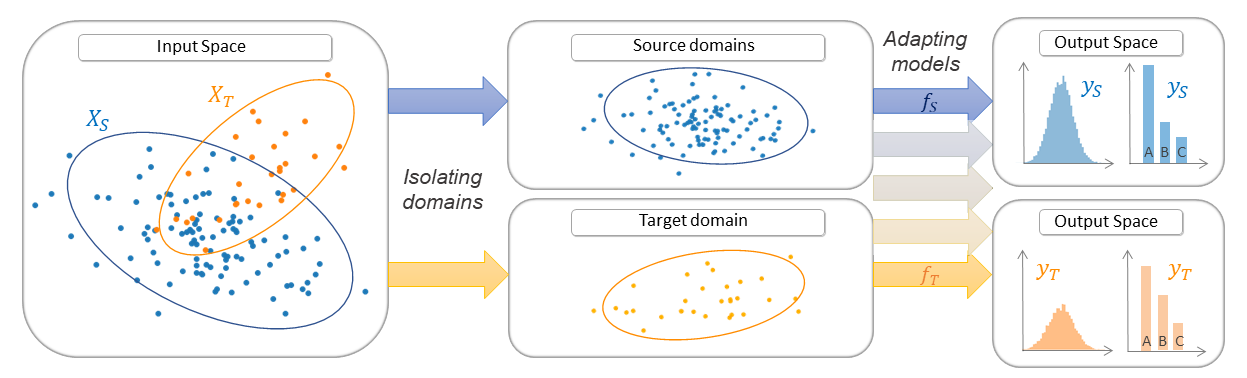}
	\caption{Parameter-based strategy illustration}
	\label{fig-param}
\end{figure}

Parameter-based methods (Figure \ref{fig-param}), also called "source-free DA", aim to adapt the parameters of a pre-trained source model to a few target observations. These DA methods are mostly used in computer vision where deep model trained on huge data sets are fine-tuned on a smaller data set of images for a specific task \citep{Oquab2014TCNN}.

\section{Installation and Usage}

\begin{figure}[ht]
	\centering
	\caption{Examples of ADAPT usage in three different settings. $X_s, y_s$ are referring to the source data and $X_t, y_t$ to the target data.}
	\; \; \, \includegraphics[width=0.9\textwidth]{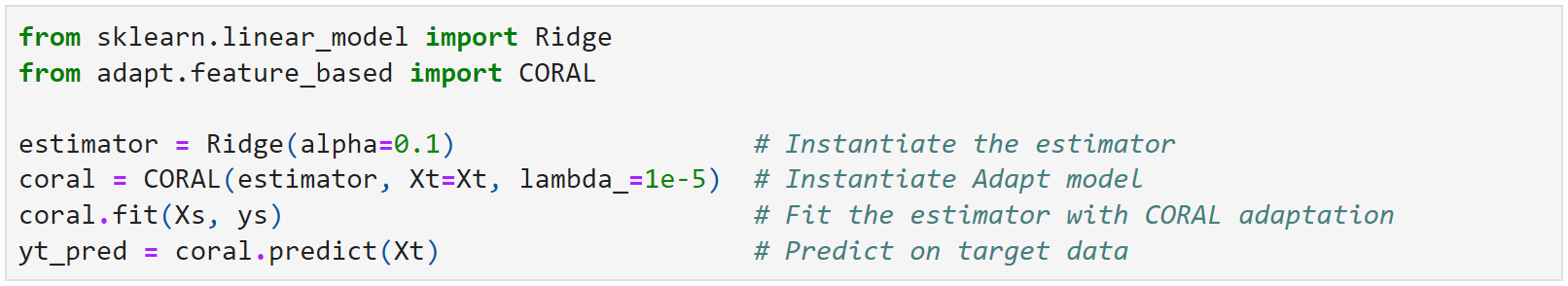}
	\; \; \,  (a) Applying CORAL \citep{Sun2015CORAL} under Unsupervised DA. \\
	\; \; \, \includegraphics[width=0.9\textwidth]{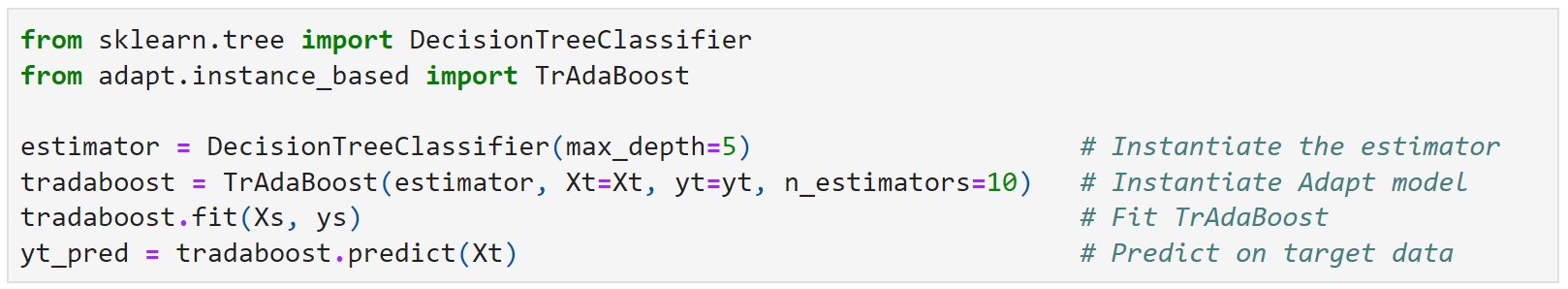}
	\; \; \, (b) Applying TrAdaBoost \citep{Dai2007TrAdaBoost} under Supervised DA. \\
	\; \; \, \includegraphics[width=0.9\textwidth]{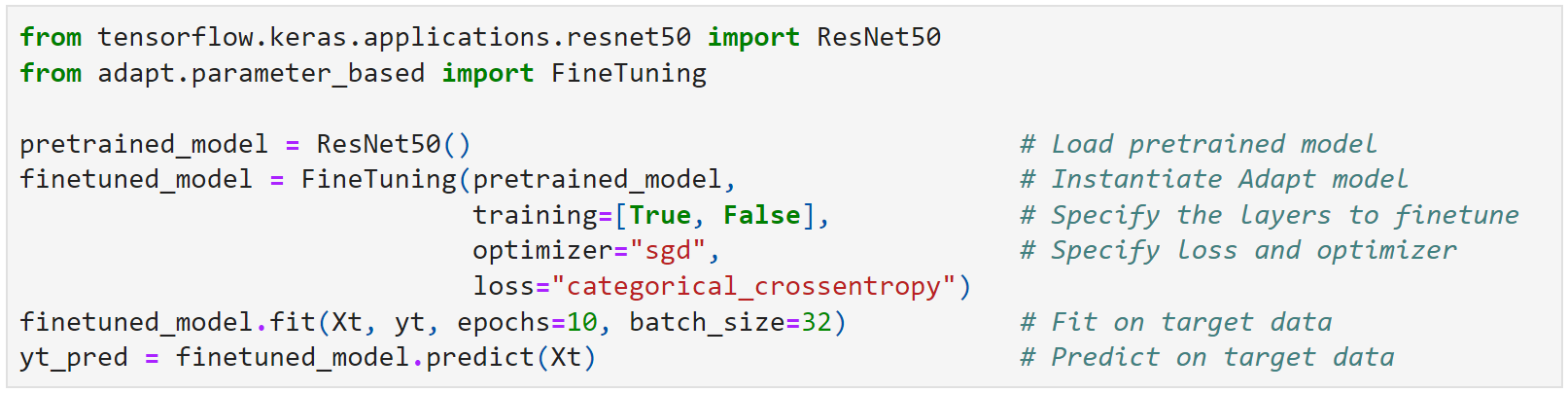}
	\; \; \, (c) Applying FineTuning \citep{Oquab2014TCNN} under Source-free DA.
	\label{fig-usage}
\end{figure}

ADAPT provides several widely used domain adaptation methods using different approaches. The provided methods allow to cope with the main DA settings encountered in real applications as \textbf{Supervised DA} and \textbf{Unsupervised DA} which respectively refer to the cases where target labels are available or not \citep{Motiian2017UDDA} (see examples in Figures \ref{fig-usage}.a, \ref{fig-usage}.b), as well as \textbf{Source-free DA} \citep{Liang2020SHOT} which is encountered when a source pre-trained model is available instead of source data (see Figure \ref{fig-usage}.c).

% \textbf{Unsupervised DA}. In this setting, the learner has access to a source labeled dataset $(X_s, y_s)$ and to an unlabeled target dataset $X_t$. The learner then seeks to use a feature transformation or importance weighting to correct the shift, for instance using CORAL \citep{Sun2015CORAL} or KMM \citep{Huang2007KMM}. (See Figure \ref{fig-usage}.a)

% \textbf{Supervised DA}. In this case, the learner has access now to additional labels in the target domain (just a few in general). He can use TrAdaBoost for example \citep{Pardoe2010boost}. (See Figure \ref{fig-usage}.b)

%\begin{figure}[ht]
%	\centering
%	\caption{Example of ADAPT usage.}
%	\includegraphics[width=\textwidth]{images/tradaboost.PNG}
%	\label{fig-usage}
%\end{figure}

% \textbf{Source-free DA}. While no source data are available but a source model is already pre-trained. Learner parameters updates using target data may be sufficient. This kind of approaches can be seen as a form of fine-tuning \cite{Oquab2014TCNN}. (See Figure \ref{fig-usage}.c)

%Finally if no source data is availabel but a pretrained source model with additional target labeled data. It can use source-free DA as Finetuning.

%\begin{figure}[ht]
%	\centering
%	\caption{Example of ADAPT usage.}
%	\includegraphics[width=\textwidth]{images/finetuning.PNG}
%	\label{fig-usage}
%\end{figure}

\section{ADAPT Guidelines}

The API is written in pure Python using scikit-learn, SciPy, NumPy, TensorFlow and cvxopt. The principal features of the API are given below:

\begin{itemize}
    \item Documentation: Each method is documented following the standards of scikit-learn. Algorithms explanations are provided along with a full description of the parameters. For each proposed method, illustrative examples are given on both synthetic datasets and real DA problems to offer visual understanding of the methods and empirical comparisons on known DA issues.
    \item Code Quality : checkers are used in all implemented objects to ensure that arguments defined by the user are valid and throw comprehensive warnings and exceptions to help the user. The code is tested with an high coverage and illustrative examples visually show that the methods are behaving as expected.
    \item Developer: ADAPT is released under a BSD2 License on GitHub. Anyone can contribute to the project by reporting issues and/or making pull requests. A Developer Guide is given to help DA researcher to include their works. Continuous integration is implemented to check the code compliance with unit tests.
\end{itemize}

\section{Conclusion and Future work}

Since its release, ADAPT has already been used for several research and industrial problems as fall detection \citep{Minvielle2019ClassImbalanceSTRUT}, tire design \citep{mathelin2021handling} and even for cosmology applications \citep{gilda2021unsupervised}. Future work will focus on adding more diversified algorithms to handle multisource and semi-supervised domain adaptation.

\newpage

\appendix

\section*{Appendix A : list of implemented methods}

\begin{table*}[ht]
	\caption{List of the implemented methods in the ADAPT library.}
	\label{table-all}
	\centering
	\begin{tabular}{c|c|c|c}
		\toprule
		& Method & Supervision & Stages \\
		\midrule
	    & FA \citep{Daume2007FAM} & SDA & 2-stages \\
        & TCA \citep{Pan2010TCA} & UDA & 2-stages \\
            & fMMD \citep{uguroglu2011fMMD} & UDA & 2-stages \\
            & SA \citep{fernando2013SA} & UDA & 2-stages \\
		\multirow{2}{*}{\begin{turn}{90}Feature\end{turn}} & CORAL \citep{Sun2015CORAL} & UDA & 2-stages \\
		& DeepCORAL \citep{Sun2016DeepCORAL} & UDA & 1-stage \\
		& DANN \citep{Ganin2016DANN} & UDA & 1-stage \\
		& ADDA \citep{Tzeng2017ADDA} & UDA & 2-stage \\
        & WDGRL \citep{Shen2017WDANN} & UDA & 1-stage \\
            & CCSA \citep{Motiian2017UDDA} & SDA & 1-stage \\
            & CDAN \citep{Long2018CDAN} & UDA & 1-stage \\
            & MCD \citep{Saito2018MCD} & UDA & 1-stage \\
            & MDD \citep{Zhang2019MDD} & UDA & 1-stage \\
		\midrule
		& KMM \citep{Huang2007KMM} & UDA & 2-stages \\
		 &KLIEP \citep{Sugiyama2007KLIEP} & UDA & 2-stages \\
  & TrAdaBoost \citep{Dai2007TrAdaBoost} & SDA & 1-stage \\
  & IWC \citep{bickel2007IWC} & UDA & 2-stages \\
		& LDM \citep{Mansour2009DATheory} & UDA & 2-stages \\ 
  & ULSIF \citep{Kanamori2009ULSIF} & UDA & 2-stages \\
		\multirow{2}{*}{\begin{turn}{90}Instance\end{turn}} & TrAdaBoostR2 \citep{Pardoe2010boost} & SDA & 1-stage \\
		&TwoStages-TrAdaBoostR2 \citep{Pardoe2010boost} & SDA & 1-stage \\
  & NNW \citep{loog2012NearestNeighborsWeighting} & UDA & 2-stages \\
  & RULSIF \citep{yamada2013RULSIF} & UDA & 2-stages \\
  & WANN \citep{deMathelin2020WANN} & SDA & 1-stage \\
  & IWN \citep{de2022IWN} & UDA & 2-stages \\
		\midrule
		& Regular Transfer LR \citep{chelba2006Adaptation} & SDA & 1-stage \\
		& Regular Transfer LC \citep{chelba2006Adaptation} & SDA & 1-stage \\
		\multirow{2}{*}{\begin{turn}{90}Param.\end{turn}} & Regular Transfer NN \citep{chelba2006Adaptation} & SDA & 1-stage \\
  & Fine-Tuning \citep{Oquab2014TCNN} & SDA & 1-stage \\
  & SER-STRUT \citep{Segev2017SERSTRUCT} & SDA & 1-stage \\
  & SER*-STRUT* \citep{Minvielle2019ClassImbalanceSTRUT} & SDA & 1-stage \\
		\bottomrule
	\end{tabular}
\end{table*}


\begin{thebibliography}{50}
\providecommand{\natexlab}[1]{#1}
\providecommand{\url}[1]{\texttt{#1}}
\expandafter\ifx\csname urlstyle\endcsname\relax
  \providecommand{\doi}[1]{doi: #1}\else
  \providecommand{\doi}{doi: \begingroup \urlstyle{rm}\Url}\fi

\bibitem[Abadi et~al.(2015)Abadi, Agarwal, Barham, Brevdo, Chen, Citro,
  Corrado, Davis, Dean, Devin, Ghemawat, Goodfellow, Harp, Irving, Isard, Jia,
  Jozefowicz, Kaiser, Kudlur, Levenberg, Man\'{e}, Monga, Moore, Murray, Olah,
  Schuster, Shlens, Steiner, Sutskever, Talwar, Tucker, Vanhoucke, Vasudevan,
  Vi\'{e}gas, Vinyals, Warden, Wattenberg, Wicke, Yu, and
  Zheng]{Abadi2015Tensorflow}
Mart\'{\i}n Abadi, Ashish Agarwal, Paul Barham, Eugene Brevdo, Zhifeng Chen,
  Craig Citro, Greg~S. Corrado, Andy Davis, Jeffrey Dean, Matthieu Devin,
  Sanjay Ghemawat, Ian Goodfellow, Andrew Harp, Geoffrey Irving, Michael Isard,
  Yangqing Jia, Rafal Jozefowicz, Lukasz Kaiser, Manjunath Kudlur, Josh
  Levenberg, Dandelion Man\'{e}, Rajat Monga, Sherry Moore, Derek Murray, Chris
  Olah, Mike Schuster, Jonathon Shlens, Benoit Steiner, Ilya Sutskever, Kunal
  Talwar, Paul Tucker, Vincent Vanhoucke, Vijay Vasudevan, Fernanda Vi\'{e}gas,
  Oriol Vinyals, Pete Warden, Martin Wattenberg, Martin Wicke, Yuan Yu, and
  Xiaoqiang Zheng.
\newblock {TensorFlow}: Large-scale machine learning on heterogeneous systems,
  2015.
\newblock URL \url{https://www.tensorflow.org/}.
\newblock Software available from tensorflow.org.

\bibitem[Bickel et~al.(2007)Bickel, Br{\"u}ckner, and Scheffer]{bickel2007IWC}
Steffen Bickel, Michael Br{\"u}ckner, and Tobias Scheffer.
\newblock Discriminative learning for differing training and test
  distributions.
\newblock In \emph{Proceedings of the 24th international conference on Machine
  learning}, pages 81--88, 2007.

\bibitem[Chelba and Acero(2006)]{chelba2006Adaptation}
Ciprian Chelba and Alex Acero.
\newblock Adaptation of maximum entropy capitalizer: Little data can help a
  lot.
\newblock \emph{Computer Speech \& Language}, 20\penalty0 (4):\penalty0
  382--399, 2006.

\bibitem[Courty et~al.(2016)Courty, Flamary, Tuia, and
  Rakotomamonjy]{Courty2016OTDA}
Nicolas Courty, R{\'e}mi Flamary, Devis Tuia, and Alain Rakotomamonjy.
\newblock Optimal transport for domain adaptation.
\newblock \emph{IEEE transactions on pattern analysis and machine
  intelligence}, 39\penalty0 (9):\penalty0 1853--1865, 2016.

\bibitem[Dai et~al.(2007)Dai, Yang, Xue, and Yu]{Dai2007TrAdaBoost}
Wenyuan Dai, Qiang Yang, Gui-Rong Xue, and Yong Yu.
\newblock Boosting for transfer learning.
\newblock In \emph{Proceedings of the 24th International Conference on Machine
  Learning}, volume 227, pages 193--200, 01 2007.
\newblock \doi{10.1145/1273496.1273521}.

\bibitem[Daum{\'e}~III(2007)]{Daume2007FAM}
Hal Daum{\'e}~III.
\newblock Frustratingly easy domain adaptation.
\newblock In \emph{Proceedings of the 45th Annual Meeting of the Association of
  Computational Linguistics}, pages 256--263, Prague, Czech Republic, June
  2007. Association for Computational Linguistics.
\newblock URL \url{https://www.aclweb.org/anthology/P07-1033}.

\bibitem[Davidson(2018)]{DDAN}
Erlend Davidson.
\newblock Deep domain adaptation networks, 2018.
\newblock URL \url{https://github.com/erlendd/ddan}.

\bibitem[de~Mathelin et~al.(2020)de~Mathelin, Richard, Deheeger, Mougeot, and
  Vayatis]{deMathelin2020WANN}
Antoine de~Mathelin, Guillaume Richard, Francois Deheeger, Mathilde Mougeot,
  and Nicolas Vayatis.
\newblock Adversarial weighting for domain adaptation in regression.
\newblock \emph{arXiv preprint arXiv:2006.08251}, 2020.

\bibitem[de~Mathelin et~al.(2021)de~Mathelin, Deheeger, Richard, Mougeot, and
  Vayatis]{Demathelin2021ADAPT}
Antoine de~Mathelin, Fran{\c{c}}ois Deheeger, Guillaume Richard, Mathilde
  Mougeot, and Nicolas Vayatis.
\newblock Adapt: Awesome domain adaptation python toolbox.
\newblock \emph{arXiv preprint arXiv:2107.03049}, 2021.

\bibitem[de~Mathelin et~al.(2022)de~Mathelin, Deheeger, Mougeot, and
  Vayatis]{de2022IWN}
Antoine de~Mathelin, Francois Deheeger, Mathilde Mougeot, and Nicolas Vayatis.
\newblock Fast and accurate importance weighting for correcting sample bias.
\newblock \emph{arXiv preprint arXiv:2209.04215}, 2022.

\bibitem[Fernando et~al.(2013)Fernando, Habrard, Sebban, and
  Tuytelaars]{fernando2013SA}
Basura Fernando, Amaury Habrard, Marc Sebban, and Tinne Tuytelaars.
\newblock Unsupervised visual domain adaptation using subspace alignment.
\newblock In \emph{Proceedings of the IEEE international conference on computer
  vision}, pages 2960--2967, 2013.

\bibitem[Ganin et~al.(2016)Ganin, Ustinova, Ajakan, Germain, Larochelle,
  Laviolette, Marchand, and Lempitsky]{Ganin2016DANN}
Yaroslav Ganin, Evgeniya Ustinova, Hana Ajakan, Pascal Germain, Hugo
  Larochelle, Fran\c{c}ois Laviolette, Mario Marchand, and Victor Lempitsky.
\newblock Domain-adversarial training of neural networks.
\newblock \emph{J. Mach. Learn. Res.}, 17\penalty0 (1):\penalty0 2096--2030,
  January 2016.
\newblock ISSN 1532-4435.

\bibitem[Gilda et~al.(2021)Gilda, de~Mathelin, Bellstedt, and
  Richard]{gilda2021unsupervised}
Sankalp Gilda, Antoine de~Mathelin, Sabine Bellstedt, and Guillaume Richard.
\newblock Unsupervised domain adaptation for constraining star formation
  histories.
\newblock \emph{arXiv preprint arXiv:2112.14072}, 2021.

\bibitem[Gulrajani and Lopez-Paz(2020)]{DomainBed}
Ishaan Gulrajani and David Lopez-Paz.
\newblock In search of lost domain generalization.
\newblock In \emph{International Conference on Learning Representations}, 2020.

\bibitem[Huang et~al.(2007)Huang, Gretton, Borgwardt, Sch\"{o}lkopf, and
  Smola]{Huang2007KMM}
Jiayuan Huang, Arthur Gretton, Karsten Borgwardt, Bernhard Sch\"{o}lkopf, and
  Alex~J. Smola.
\newblock Correcting sample selection bias by unlabeled data.
\newblock In B.~Sch\"{o}lkopf, J.~C. Platt, and T.~Hoffman, editors,
  \emph{Advances in Neural Information Processing Systems 19}, pages 601--608.
  MIT Press, 2007.

\bibitem[Junguang~Jiang(2020)]{Jiang2020dalib}
Mingsheng~Long Junguang~Jiang, Bo~Fu.
\newblock Transfer-learning-library.
\newblock \url{https://github.com/thuml/Transfer-Learning-Library}, 2020.

\bibitem[Kanamori et~al.(2009)Kanamori, Hido, and Sugiyama]{Kanamori2009ULSIF}
Takafumi Kanamori, Shohei Hido, and Masashi Sugiyama.
\newblock A least-squares approach to direct importance estimation.
\newblock \emph{The Journal of Machine Learning Research}, 10:\penalty0
  1391--1445, 2009.

\bibitem[Liang et~al.(2020)Liang, Hu, and Feng]{Liang2020SHOT}
Jian Liang, Dapeng Hu, and Jiashi Feng.
\newblock Do we really need to access the source data? source hypothesis
  transfer for unsupervised domain adaptation.
\newblock In \emph{International Conference on Machine Learning}, pages
  6028--6039. PMLR, 2020.

\bibitem[Long et~al.(2018)Long, CAO, Wang, and Jordan]{Long2018CDAN}
Mingsheng Long, ZHANGJIE CAO, Jianmin Wang, and Michael~I Jordan.
\newblock Conditional adversarial domain adaptation.
\newblock In S.~Bengio, H.~Wallach, H.~Larochelle, K.~Grauman, N.~Cesa-Bianchi,
  and R.~Garnett, editors, \emph{Advances in Neural Information Processing
  Systems 31}, pages 1640--1650. Curran Associates, Inc., 2018.

\bibitem[Loog(2012)]{loog2012NearestNeighborsWeighting}
Marco Loog.
\newblock Nearest neighbor-based importance weighting.
\newblock In \emph{2012 IEEE International Workshop on Machine Learning for
  Signal Processing}, pages 1--6. IEEE, 2012.

\bibitem[Mansour et~al.(2009)Mansour, Mohri, and
  Rostamizadeh]{Mansour2009DATheory}
Yishay Mansour, Mehryar Mohri, and Afshin Rostamizadeh.
\newblock Domain adaptation: Learning bounds and algorithms.
\newblock In \emph{COLT}, 2009.

\bibitem[Mathelin et~al.(2021)Mathelin, Deheeger, Mougeot, and
  Vayatis]{mathelin2021handling}
Antoine~De Mathelin, Fran{\c{c}}ois Deheeger, Mathilde Mougeot, and Nicolas
  Vayatis.
\newblock Handling distribution shift in tire design.
\newblock In \emph{NeurIPS 2021 Workshop on Distribution Shifts: Connecting
  Methods and Applications}, 2021.
\newblock URL \url{https://openreview.net/forum?id=W0fKtUQgcRR}.

\bibitem[Minvielle et~al.(2019)Minvielle, Atiq, Peignier, and
  Mougeot]{Minvielle2019ClassImbalanceSTRUT}
Ludovic Minvielle, Mounir Atiq, Sergio Peignier, and Mathilde Mougeot.
\newblock Transfer learning on decision tree with class imbalance.
\newblock In \emph{2019 IEEE 31st International Conference on Tools with
  Artificial Intelligence (ICTAI)}, pages 1003--1010, 2019.
\newblock \doi{10.1109/ICTAI.2019.00141}.

\bibitem[Motiian et~al.(2017)Motiian, Piccirilli, Adjeroh, and
  Doretto]{Motiian2017UDDA}
Saeid Motiian, Marco Piccirilli, Donald~A Adjeroh, and Gianfranco Doretto.
\newblock Unified deep supervised domain adaptation and generalization.
\newblock In \emph{Proceedings of the IEEE International Conference on Computer
  Vision}, pages 5715--5725, 2017.

\bibitem[Musgrave(2021)]{PytorchAdapt}
Kevin Musgrave.
\newblock Pytorch-adapt.
\newblock \url{https://github.com/KevinMusgrave/pytorch-adapt}, 2021.

\bibitem[Oquab et~al.(2014)Oquab, Bottou, Laptev, and Sivic]{Oquab2014TCNN}
M.~Oquab, L.~Bottou, I.~Laptev, and J.~Sivic.
\newblock Learning and transferring mid-level image representations using
  convolutional neural networks.
\newblock In \emph{CVPR}, 2014.

\bibitem[Pan et~al.(2010)Pan, Tsang, Kwok, and Yang]{Pan2010TCA}
Sinno~Jialin Pan, Ivor~W Tsang, James~T Kwok, and Qiang Yang.
\newblock Domain adaptation via transfer component analysis.
\newblock \emph{IEEE Transactions on Neural Networks}, 22\penalty0
  (2):\penalty0 199--210, 2010.

\bibitem[Pardoe and Stone(2010)]{Pardoe2010boost}
David Pardoe and Peter Stone.
\newblock Boosting for regression transfer.
\newblock In \emph{Proceedings of the 27th International Conference on Machine
  Learning (ICML)}, June 2010.

\bibitem[Paszke et~al.(2019)Paszke, Gross, Massa, Lerer, Bradbury, Chanan,
  Killeen, Lin, Gimelshein, Antiga, Desmaison, Kopf, Yang, DeVito, Raison,
  Tejani, Chilamkurthy, Steiner, Fang, Bai, and Chintala]{Paszke2019Pytorch}
Adam Paszke, Sam Gross, Francisco Massa, Adam Lerer, James Bradbury, Gregory
  Chanan, Trevor Killeen, Zeming Lin, Natalia Gimelshein, Luca Antiga, Alban
  Desmaison, Andreas Kopf, Edward Yang, Zachary DeVito, Martin Raison, Alykhan
  Tejani, Sasank Chilamkurthy, Benoit Steiner, Lu~Fang, Junjie Bai, and Soumith
  Chintala.
\newblock Pytorch: An imperative style, high-performance deep learning library.
\newblock In H.~Wallach, H.~Larochelle, A.~Beygelzimer, F.~d\textquotesingle
  Alch\'{e}-Buc, E.~Fox, and R.~Garnett, editors, \emph{Advances in Neural
  Information Processing Systems 32}, pages 8024--8035. Curran Associates,
  Inc., 2019.
\newblock URL
  \url{http://papers.neurips.cc/paper/9015-pytorch-an-imperative-style-high-performance-deep-learning-library.pdf}.

\bibitem[Pedregosa et~al.(2011)Pedregosa, Varoquaux, Gramfort, Michel, Thirion,
  Grisel, Blondel, Prettenhofer, Weiss, Dubourg, Vanderplas, Passos,
  Cournapeau, Brucher, Perrot, and Duchesnay]{Pedregosa2011scikit-learn}
F.~Pedregosa, G.~Varoquaux, A.~Gramfort, V.~Michel, B.~Thirion, O.~Grisel,
  M.~Blondel, P.~Prettenhofer, R.~Weiss, V.~Dubourg, J.~Vanderplas, A.~Passos,
  D.~Cournapeau, M.~Brucher, M.~Perrot, and E.~Duchesnay.
\newblock Scikit-learn: Machine learning in {P}ython.
\newblock \emph{Journal of Machine Learning Research}, 12:\penalty0 2825--2830,
  2011.

\bibitem[Ravishankar et~al.(2016)Ravishankar, Sudhakar, Venkataramani,
  Thiruvenkadam, Annangi, Babu, and Vaidya]{Ravishankar2016TLMedicalImages}
Hariharan Ravishankar, Prasad Sudhakar, Rahul Venkataramani, Sheshadri
  Thiruvenkadam, Pavan Annangi, Narayanan Babu, and Vivek Vaidya.
\newblock Understanding the mechanisms of deep transfer learning for medical
  images.
\newblock In \emph{Deep learning and data labeling for medical applications},
  pages 188--196. Springer, 2016.

\bibitem[Saito et~al.(2018)Saito, Watanabe, Ushiku, and Harada]{Saito2018MCD}
Kuniaki Saito, Kohei Watanabe, Yoshitaka Ushiku, and Tatsuya Harada.
\newblock Maximum classifier discrepancy for unsupervised domain adaptation.
\newblock In \emph{Proceedings of the IEEE Conference on Computer Vision and
  Pattern Recognition}, pages 3723--3732, 2018.

\bibitem[Schneider et~al.(2018)Schneider, Ecker, Macke, and
  Bethge]{Schneider2018salad}
Steffen Schneider, Alexander~S. Ecker, Jakob~H. Macke, and Matthias Bethge.
\newblock Salad: A toolbox for semi-supervised adaptive learning across
  domains, 2018.
\newblock URL \url{https://openreview.net/forum?id=S1lTifykqm}.

\bibitem[{Segev} et~al.(2017){Segev}, {Harel}, {Mannor}, {Crammer}, and
  {El-Yaniv}]{Segev2017SERSTRUCT}
N.~{Segev}, M.~{Harel}, S.~{Mannor}, K.~{Crammer}, and R.~{El-Yaniv}.
\newblock Learn on source, refine on target: A model transfer learning
  framework with random forests.
\newblock \emph{IEEE Transactions on Pattern Analysis and Machine
  Intelligence}, 39\penalty0 (9):\penalty0 1811--1824, 2017.

\bibitem[Shen et~al.(2018)Shen, Qu, Zhang, and Yu]{Shen2017WDANN}
Jian Shen, Yanru Qu, Weinan Zhang, and Yong Yu.
\newblock Wasserstein distance guided representation learning for domain
  adaptation.
\newblock In \emph{Proceedings of the AAAI Conference on Artificial
  Intelligence}, volume~32, 2018.

\bibitem[Sugiyama et~al.(2007)Sugiyama, Nakajima, Kashima, B\"{u}nau, and
  Kawanabe]{Sugiyama2007KLIEP}
Masashi Sugiyama, Shinichi Nakajima, Hisashi Kashima, Paul~von B\"{u}nau, and
  Motoaki Kawanabe.
\newblock Direct importance estimation with model selection and its application
  to covariate shift adaptation.
\newblock In \emph{Proceedings of the 20th International Conference on Neural
  Information Processing Systems}, NIPS’07, page 1433–1440, Red Hook, NY,
  USA, 2007. Curran Associates Inc.
\newblock ISBN 9781605603520.

\bibitem[Sun and Saenko(2016)]{Sun2016DeepCORAL}
Baochen Sun and Kate Saenko.
\newblock Deep coral: Correlation alignment for deep domain adaptation.
\newblock In \emph{European conference on computer vision}, pages 443--450.
  Springer, 2016.

\bibitem[Sun et~al.(2016)Sun, Feng, and Saenko]{Sun2015CORAL}
Baochen Sun, Jiashi Feng, and Kate Saenko.
\newblock Return of frustratingly easy domain adaptation.
\newblock In \emph{Proceedings of the AAAI Conference on Artificial
  Intelligence}, volume~30, 2016.

\bibitem[Tousch and Renaudin(2020)]{pytorchada}
Anne-Marie Tousch and Christophe Renaudin.
\newblock (yet) another domain adaptation library, 2020.
\newblock URL \url{https://github.com/criteo-research/pytorch-ada}.

\bibitem[Tzeng et~al.(2017)Tzeng, Hoffman, Saenko, and Darrell]{Tzeng2017ADDA}
Eric Tzeng, Judy Hoffman, Kate Saenko, and Trevor Darrell.
\newblock Adversarial discriminative domain adaptation.
\newblock In \emph{Proceedings of the IEEE Conference on Computer Vision and
  Pattern Recognition}, pages 7167--7176, 2017.

\bibitem[Uguroglu and Carbonell(2011)]{uguroglu2011fMMD}
Selen Uguroglu and Jaime Carbonell.
\newblock Feature selection for transfer learning.
\newblock In \emph{Joint European Conference on Machine Learning and Knowledge
  Discovery in Databases}, pages 430--442. Springer, 2011.

\bibitem[Vercruyssen(2020)]{transfertools}
V~Vercruyssen.
\newblock Transfertools, 2020.
\newblock URL \url{https://github.com/Vincent-Vercruyssen/transfertools}.

\bibitem[Weiss et~al.(2016)Weiss, Khoshgoftaar, and Wang]{Weiss2016survey}
Karl Weiss, Taghi~M. Khoshgoftaar, and DingDing Wang.
\newblock A survey of transfer learning.
\newblock \emph{Journal of Big Data}, 3\penalty0 (1):\penalty0 9, May 2016.
\newblock ISSN 2196-1115.
\newblock \doi{10.1186/s40537-016-0043-6}.
\newblock URL \url{https://doi.org/10.1186/s40537-016-0043-6}.

\bibitem[Wouter(2015)]{libTLDA}
Kouw Wouter.
\newblock libtlda.
\newblock \url{https://github.com/wmkouw/libTLDA}, 2015.

\bibitem[Yamada et~al.(2013)Yamada, Suzuki, Kanamori, Hachiya, and
  Sugiyama]{yamada2013RULSIF}
Makoto Yamada, Taiji Suzuki, Takafumi Kanamori, Hirotaka Hachiya, and Masashi
  Sugiyama.
\newblock Relative density-ratio estimation for robust distribution comparison.
\newblock \emph{Neural computation}, 25\penalty0 (5):\penalty0 1324--1370,
  2013.

\bibitem[Yan(2016)]{DAtoolbox}
Ke~Yan.
\newblock Domain adaptation toolbox, 2016.
\newblock URL \url{https://github.com/viggin/domain-adaptation-toolbox}.

\bibitem[Zhang et~al.(2019)Zhang, Liu, Long, and Jordan]{Zhang2019MDD}
Yuchen Zhang, Tianle Liu, Mingsheng Long, and Michael Jordan.
\newblock Bridging theory and algorithm for domain adaptation.
\newblock In Kamalika Chaudhuri and Ruslan Salakhutdinov, editors,
  \emph{Proceedings of the 36th International Conference on Machine Learning},
  volume~97 of \emph{Proceedings of Machine Learning Research}, pages
  7404--7413, Long Beach, California, USA, 09--15 Jun 2019. PMLR.

\bibitem[Zhou et~al.(2021)Zhou, Yang, Qiao, and Xiang]{Dassl}
Kaiyang Zhou, Yongxin Yang, Yu~Qiao, and Tao Xiang.
\newblock Domain adaptive ensemble learning.
\newblock \emph{IEEE Transactions on Image Processing (TIP)}, 2021.

\bibitem[Zhu et~al.(2019)Zhu, Zhuang, and Wang]{deepTL}
Yongchun Zhu, Fuzhen Zhuang, and Deqing Wang.
\newblock Aligning domain-specific distribution and classifier for cross-domain
  classification from multiple sources.
\newblock In \emph{Proceedings of the AAAI Conference on Artificial
  Intelligence}, volume~33, pages 5989--5996, 2019.

\bibitem[Zhuang et~al.(2019)Zhuang, Duan, Guo, Zhu, Xi, Qi, and
  He]{Zhuang2019TransferToolkit}
Fuzhen Zhuang, Keyu Duan, Tongjia Guo, Yongchun Zhu, Dongbo Xi, Zhiyuan Qi, and
  Qing He.
\newblock Transfer learning toolkit: Primers and benchmarks, 2019.

\end{thebibliography}
\end{document}